%% file: main.tex
\documentclass[runningheads]{llncs}

\usepackage{eccv}

\input{preamble}
\input{acronyms}

\input{paper_macros}

\usepackage{hyperref}

\usepackage{orcidlink}
\usepackage{threeparttable}

\begin{document}

\title{Unveiling Transferability in Trajectory Prediction via Latent Scene Embeddings} 

\titlerunning{Unveiling Transferability in Trajectory Prediction}

\author{Theodor Westny\inst{1}\orcidlink{0000-0001-9075-7477} \and
David Axelsson\inst{1}\orcidlink{0009-0004-1105-2110} \and
Björn Olofsson\inst{1,2}\orcidlink{0000-0003-1320-032X} \and
Erik Frisk\inst{1}\orcidlink{0000-0001-7349-1937}}

\authorrunning{T.~Westny et al.}

\institute{
Division of Vehicular Systems, Linköping University,
Linköping, Sweden\\
\email{
\{theodor.westny,
david.axelsson,
erik.frisk\}@liu.se
}
\and
Department of Automatic Control, Lund University,
Lund, Sweden\\
\email{bjorn.olofsson@control.lth.se}
}

\maketitle

\input{sec/abstract}

\input{sec/introduction}
\input{sec/related_work.tex}

\input{sec/problem_definition.tex}
\input{sec/latent_model.tex}
\input{sec/results.tex}

\input{sec/conclusion.tex}

\section*{Acknowledgments}
This research was supported by the Strategic Research Area at Linköping-Lund in Information Technology (ELLIIT) and the Wallenberg AI, Autonomous Systems and Software Program (WASP) funded by the Knut and Alice Wallenberg Foundation.
Computations were enabled by the Berzelius resource provided by the Knut and Alice Wallenberg Foundation at the National Supercomputer Centre.

\bibliographystyle{splncs04}
\bibliography{IEEEabrv,references}
\include{sec/suppl}

\end{document}

%% file: preamble.tex
\usepackage{microtype}      %
\usepackage{textcomp}       %
\usepackage{xcolor}         %
\usepackage{url}            %

\usepackage{amsmath}
\usepackage{amssymb}
\usepackage{amsfonts}
\usepackage{bm}             %
\usepackage{mathtools}      %
\usepackage{cancel}         %

\usepackage{booktabs}       %
\usepackage{makecell}       %
\usepackage{array}          %
\usepackage{siunitx}        %
\usepackage{multirow}

\usepackage{graphicx}       %
\usepackage{wrapfig}        %
\usepackage{float}          %

\usepackage{algorithm}      %
\usepackage{algorithmic}    %

\usepackage{enumitem}       %
\usepackage{pifont}         %
\usepackage{verbatim}       %

\usepackage[acronym]{glossaries} %
\usepackage{stfloats}       %
\usepackage[accsupp]{axessibility} %

\usepackage{pdflscape}  %

\definecolor{cvprblue}{rgb}{0.21,0.49,0.74}

\usepackage[breaklinks,colorlinks,allcolors=cvprblue]{hyperref}

\definecolor{r1}{HTML}{DA731D}
\definecolor{r2}{HTML}{1D90DA}
\definecolor{r3}{HTML}{19BD61}

\usepackage{orcidlink} %

\usepackage{eccvabbrv}

\usepackage{todonotes} %

%% file: acronyms.tex
\newacronym{GNN}{GNN}{graph neural network}
\newacronym{CNN}{CNN}{convolutional neural network}
\newacronym{MDN}{MDN}{mixture density network}
\newacronym{CVAE}{CVAE}{conditional variational autoencoder}
\newacronym{GAN}{GAN}{generative adversarial network}

\newacronym{MHA}{MHA}{multi-head attention}
\newacronym{GGRU}{Graph-GRU}{graph-gated recurrent unit}
\newacronym{GRU}{GRU}{gated recurrent unit}

\newacronym{HD}{HD}{high-definition}

\newacronym{ADE}{ADE}{Average Displacement Error}
\newacronym{FDE}{FDE}{Final Displacement Error}
\newacronym{MR}{MR}{Miss Rate}
\newacronym{MMD}{MMD}{maximum mean discrepancy}

\newacronym{GCN}{GCN}{graph convolutional network}
\newacronym{GAT}{GAT}{graph attention network}

\newacronym{KL}{KL}{Kullback--Leibler}
\newacronym{PCA}{PCA}{Principal Component Analysis}

%% file: paper_macros.tex
\newcommand{\cmark}{\ding{51}}%
\newcommand{\xmark}{\ding{55}}%

\definecolor{MyGreen}{HTML}{10A23B}
\definecolor{MyRed}{HTML}{DD1111}

\newcommand{\dataset}{\mathcal{D}}
\newcommand{\lanegraph}{\mathbf{M}}
\newcommand{\ifeats}{\mathbf{X}}
\newcommand{\ofeats}{\mathbf{Y}}
\newcommand{\latentmat}{\mathbf{Z}}
\newcommand{\latentscene}{\mathbf{Z}_\mathcal{S}}
\newcommand{\latentdata}{\mathbf{Z}_\mathcal{D}}

\newcommand{\latent}{\mathbf{z}}

\newcommand{\kl}{D_{\text{KL}}}

\DeclareMathOperator{\encoder}{\mathsf{Enc}}
\DeclareMathOperator{\recHead}{\mathsf{Rec}}
\DeclareMathOperator{\predHead}{\mathsf{Pred}}

\newcommand{\graph}{\mathcal{G}}
\newcommand{\node}{\mathcal{V}}
\newcommand{\edge}{\mathcal{E}}

\newcommand{\neigh}[1]{\mathbf{N}(#1)}

\newcommand{\transpose}{\top} %

\newcommand{\sigmoid}{\sigma}

%% file: sec/abstract.tex
\begin{abstract}
    The growing availability of trajectory datasets has fueled major advances in data-driven motion prediction. Yet,
    models trained on one dataset often fail to generalize beyond their training domain as a result of differences in scene
    layouts, agent behaviors, and sensing conditions. A framework that learns latent representations of
    datasets and quantifies their similarity using distributional metrics is presented. This large-scale study covers 24 major
    datasets, including the most widely used motion-prediction benchmarks, and shows that the resulting
    transferability scores strongly correlate with cross-dataset model performance. The results provide practical
    guidance for dataset selection, pretraining, and large-scale foundation models for motion prediction, paving the way
    toward more generalizable and robust predictive systems.
\end{abstract}

%% file: sec/introduction.tex
\section{Introduction}
Trajectory prediction plays an important role in applications such as autonomous driving, robotics, and human behavior analysis~\cite{huang2022survey, fang2024behavioral,wang2025deployable}. It involves forecasting the future states of agents based on past observations and interactions with their surroundings.
The topic has gained significant attention in recent years, driven by advances in deep-learning methods, increased availability of trajectory datasets~\cite{i80v2,us101v2,lerner2007crowds,pellegrini2009you,ma2019trafficpredict,chang2019argoverse,caesar2020nuscenes,houston2021one,ettinger2021large,argoverse2dataset,highDdataset,zhan2019interaction,inDdataset,rounDdataset,breuer2020opendd,exiDdataset,sinDdataset,zhang2023ad4che,uniDdataset,a43dataset,boekema2024vodp}, and open-source tools~\cite{ivanovic2023trajdata, li2023scenarionet, feng2024unitraj, westny2025toward}.
However, this progress has also highlighted an important challenge: trajectory datasets differ significantly in terms of agent types, environmental complexity, and sensing modalities.
This diversity offers valuable opportunities for comprehensive model development, but it also poses difficulties for transferring models across different domains.
A key open question is how well a model trained on one dataset generalizes to another---and whether such transferability can be predicted.

Advances in large foundation models demonstrate that training on diverse datasets can yield highly transferable
representations~\cite{bommasani2021opportunities,gao2024survey,zhou2024vision,wang2025deployable}.
Theoretically, this could be applied to trajectory prediction by pooling all available data.
However, several factors limit the practicality of this approach. 
First, real-time systems such as autonomous vehicles operate under strict computational and latency constraints, making it impractical to deploy models at the scale and with the inference costs of current foundation models.
Furthermore, while model compression and distillation techniques alleviate these issues~\cite{hinton2015distilling,gou2021knowledge,feng2024road,hegde2025distilling}, recent work shows that more data are not always better; low-quality or irrelevant data can degrade performance~\cite{lee2021deduplicating,touvron2023llama,penedo2024fineweb,zhou2025smartpretrain}. 
These observations motivate prioritizing the most relevant datasets for pretraining, rather than indiscriminately aggregating all available data.

A natural first step toward addressing this challenge is to compare datasets based on their surface-level features~\cite{amirian2020opentraj,rudenko2022atlas,ivanovic2023trajdata}.
These analyses can highlight differences in statistical properties, including agent densities, speed profiles, and map structures. 
However, they do not inherently capture emergent behavioral characteristics, such as interaction patterns, social norms, or task intent, which are defining features of multi-agent motion-prediction problems~\cite{tolstaya2021identifying}.

\begin{figure}[t]
	\centering
	\includegraphics[width=0.52\columnwidth]{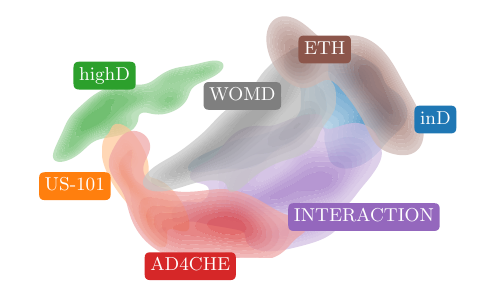}
	\caption{t-SNE projection of learned scenario embeddings (contours indicate density). Dataset proximity (\eg, ETH/inD, INTERACTION/WOMD) indicates potential for cross-dataset pretraining or knowledge transfer.}
	\vspace{-0.3cm}
	\label{fig:intro-fig}
\end{figure}

In this work, transferability is studied through a learned latent space that captures behavioral characteristics across datasets, as illustrated in \cref{fig:intro-fig}. 
This space is obtained by jointly training a single embedding model on all datasets under investigation to project scenarios into a representation where distances reflect structural and behavioral patterns in addition to surface-level feature similarities.
This learned representation provides a more meaningful basis for assessing dataset transferability by leveraging the capacity of deep models to capture complex interactions. 
Once trained, the embedding model enables querying at the dataset level, supporting data-driven recommendations of suitable source datasets for pretraining or adaptation.
In practice, transferability is estimated using the \gls{KL} divergence between Gaussian approximations of the latent dataset distributions. 
It is demonstrated that smaller latent distribution divergences correspond to improved transfer performance.

\subsection{Contributions}
The primary contributions of this paper are:
\begin{itemize}[label=--]

\item \textbf{Dataset-level representation learning for trajectory prediction.}
A latent embedding framework is introduced, modeling trajectory datasets as distributions in a shared latent space to enable comparison through their distributional geometry.

\item \textbf{A directional probabilistic measure of transferability.}
A divergence-based transferability metric is proposed to capture asymmetric transfer success, demonstrating that latent distances predict both zero-shot and fine-tuned performance. %

\item \textbf{A systematic large-scale study of cross-dataset transfer.}
A comprehensive evaluation across 24 trajectory datasets and 552 transfer pairs is conducted, revealing underutilized datasets that improve performance on popular motion-prediction benchmarks.

\end{itemize}
Implementations are available at \url{https://github.com/westny/transferatlas}.

%% file: sec/related_work.tex
\section{Related work}
\label{sec:related_work}

\subsection{Trajectory prediction}
Trajectory prediction remains an active area of research as a result of its central role in domains such as autonomous driving and robotics~\cite{wang2025deployable,huang2022survey,fang2024behavioral}.
Advances in deep learning and increased data availability have made learning-based methods the dominant approach.
A wide range of architectures have been explored, including recurrent networks~\cite{alahi2016social,deo2018convolutional,messaoud2020attention,westny2023eval}, convolutional encoders~\cite{deo2018convolutional,cui2019multimodal,zhao2019multi}, Transformer-based models~\cite{giuliari2021transformer,liu2021multimodal,huang2022multi,mao2023leapfrog}, and \glspl{GNN}~\cite{diehl2019graph,li2019grip,salzmann2020trajectron,li2021spatio,gilles2022gohome,westny2023mtp,wang2023spatio}.

A substantial portion of the literature focuses on modeling interactivity, \ie, how agents influence one another and how they are affected by their surrounding environment.
This includes pooling-based methods~\cite{alahi2016social,gupta2018social,deo2018convolutional,zhao2019multi,messaoud2020attention}, as well as approaches that explicitly model interactions using \glspl{GNN} or Transformer-based architectures~\cite{diehl2019graph,salzmann2020trajectron,li2021spatio,yuan2021agentformer,gilles2022gohome,westny2023mtp,mao2023leapfrog}.
Environmental context is processed using specialized modules and usually incorporated through either rasterized scene encodings~\cite{cui2019multimodal,marchetti2020mantra,messaoud2021trajectory,zhao2019multi,park2020diverse,salzmann2020trajectron,li2021spatio,zhou2023query} or as vectorized maps~\cite{liang2020learning,gao2023dynamic,deo2022multimodal,gilles2022gohome,zhou2023query}. %

Further, probabilistic modeling is widely used to capture the inherent uncertainty of future behavior. 
These models can be broadly categorized as \emph{discriminative}, which directly predict multi-modal trajectory distributions~\cite{hu2018probabilistic,deo2018convolutional,messaoud2020attention,messaoud2021trajectory,westny2023mtp,zhou2023query,liu2023tracing}, or \emph{generative}, which aim to learn the underlying distribution from which future trajectories can be sampled~\cite{gupta2018social,amirian2019social,sadeghian2019sophie,zhao2019multi,salzmann2020trajectron,yuan2021agentformer,li2021spatio,xu2022socialvae,gu2022stochastic,jiang2023motiondiffuser,mao2023leapfrog,westny2024diffusion,bae2024singulartrajectory,choi2024dice}.

\subsection{Domain adaptation and generalization}
Improving model robustness across environments often involves either domain adaptation or domain generalization. 
Domain adaptation aims to reduce the discrepancy between the source and target data distributions, while domain generalization seeks models that perform well on unseen domains without additional tuning. In the field of trajectory prediction, several works explore these themes.

Recent approaches propose architectural adaptations such as ensemble-based methods~\cite{westny2021vehicle}, or specialized modules~\cite{wang2023bridging,zhang2024spatial} to improve generalization.
Other strategies tackle this problem by learning shared representations across datasets~\cite{shi2023metatraj,bae2024singulartrajectory,park2024improving}, or by designing dataset-agnostic input abstractions that reduce reliance on domain-specific cues~\cite{jaipuria2018curbside,hu2022scenario,gilles2022uncertainty,ye2023improving,ivanovic2023trajdata,feng2024unitraj,zhou2025smartpretrain}.

On the domain adaptation side, recent work proposes techniques such as attention-based adaptation~\cite{xu2022adaptive}, feature alignment~\cite{geng2023adaptive}, and instance-level augmentations~\cite{kong2024adaptive}. Other approaches adapt pre-trained models to new domains~\cite{ullrich2024transfer} or introduce meta-learning strategies for online or few-shot adaptation~\cite{ivanovic2023expanding}.

While domain adaptation and generalization are central themes in this work, the focus is not on developing new methods
to improve these properties, but rather on understanding whether they can be predicted. Furthermore, evaluations are
typically restricted to a limited set of source--target pairs. This leaves a clear gap for research that investigates
transferability across a broader, more diverse spectrum of datasets.

\subsection{Dataset similarity and transferability metrics}
Quantifying dataset similarity is important for anticipating transfer learning success~\cite{yosinski2014transferable,wang2019characterizing,achille2019task2vec,clark2022predicting,ehrig2024impact}.
A closely related approach to ours is presented in~\cite{clark2022predicting}, where methods to estimate transferability between time-series datasets are proposed. 
A shared autoencoder embeds source and target data into a common latent space, from which metrics such as the L1 distance between dataset centroids are derived and shown to correlate with transfer learning gains.
In contrast, this work focuses on trajectory prediction, where spatio-temporal structure, interaction dynamics, and
map-based context require specialized modeling~\cite{wang2025deployable,huang2022survey,fang2024behavioral}. In addition, probabilistic and non-symmetric measures, such as
\gls{KL} divergence, are considered to capture the inherently directional nature of transferability.

Related work on latent representations of trajectory datasets includes ScenarioNet~\cite{li2023scenarionet}, which utilizes TrafficGen~\cite{feng2023trafficgen} to obtain scenario embeddings that are then projected onto a lower-dimensional manifold. These projections highlight differences between synthetic and real data as well as residual variation across datasets. While primarily used for visualization, such an approach also highlights the potential of using latent space analyses to identify distributional gaps.

%% file: sec/problem_definition.tex
\section{Problem definition}
\label{sec:problem-definition}
The problem of learning the approximate distributions of multiple trajectory datasets and using these distributions to quantify pairwise transferability is considered.
Each dataset consists of interactive \emph{scenes}, where a scene is defined as a set of agents (\eg, vehicles, pedestrians) and their associated features (\eg, position, velocity) that evolves over time.
The temporal span of a scene is divided into a historical segment and a future segment.
Each time instant in the scene is modeled as an undirected graph $\graph = (\node, \edge)$, where $\node$ is the set of agents (nodes) and $\edge$ is the set of edges representing the interactions between said agents.
Let $N = |\node|$ be the number of agents, $H$ the length of the historical segment, and $F$ the length of the future segment. The node input features are summarized in a tensor $\ifeats \in \mathbb{R}^{N \times H \times D}$, and the output features in $\ofeats \in \mathbb{R}^{N \times F \times D}$, where $D=5$, corresponding to 2D position, 2D velocity, and orientation.
When available, the map $\lanegraph$ is also incorporated, encoded as a lane graph with associated features, shared across all time instants of the scene.
For brevity, the notation is overloaded, letting $\ifeats$ implicitly denote the pair $(\ifeats, \lanegraph)$ whenever map information is available.

Next, latent variables $\latent_i$ are introduced for each node $i$, summarized in a matrix $\latentmat \in \mathbb{R}^{N \times L}$.
Finally, for a scene $\mathcal{S}$, the latent scene embedding $\latentscene \in \mathbb{R}^{L}$ is defined, and for a dataset $\mathcal{D}$, the corresponding latent dataset embedding $\latentdata \in \mathbb{R}^{L}$, where $L$ denotes the latent dimension. Unless otherwise stated, $L=32$ in this work.
The objective is two-fold:
\begin{enumerate}
    \item Learn the latent dataset-level representations $\{{\latentdata}_j\}_{j=1}^K$ for datasets\\ $\dataset_1, \dataset_2, \ldots, \dataset_K$ and use them to quantify pairwise transferability based on distributional (dis)similarity.

    \item Empirically validate the inferred transferability by training a trajectory prediction model \( \ifeats \mapsto \ofeats \) on dataset \( \dataset_i \) and evaluating it on a different dataset \( \dataset_j \), for multiple pairs \( (\dataset_i, \dataset_j) \).

\end{enumerate}

%% file: sec/latent_model.tex
\begin{figure*}[t]
	\centering
	\includegraphics[width=1\textwidth]{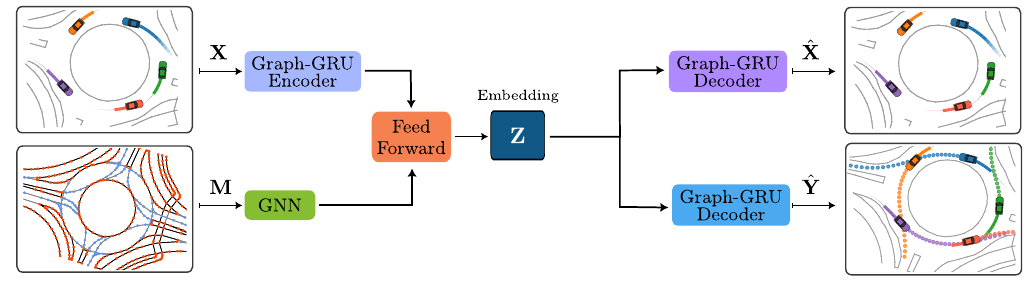}
	\caption{Latent embedding model architecture. An encoder maps input features to node-level latent variables, which are aggregated into scene- and dataset-level embeddings. Two decoder heads handle feature reconstruction (predicting $\ifeats$) and future state forecasting (predicting $\ofeats$). Training optimizes a combined reconstruction and forecasting loss.}
    \label{fig:model-architecture}
\end{figure*}

\section{Latent embedding model}
\label{sec:latent-model}
To construct compact and comparable representations of the datasets under consideration, a deep latent embedding model is employed, designed to learn expressive codes of the individual samples (scenes) comprising each dataset. 

To learn representations that capture all aspects of the data under a setting relevant to trajectory prediction, the model is trained using two complementary supervision signals: reconstruction of observed inputs and prediction of future states. The predictive objective reflects the trajectory forecasting setting, where past motion is observed and future motion is unknown, while reconstruction promotes information preservation. %
Latent regularization is used to encourage informative and stable representations~\cite[p.~564]{bishop2024deep}. In particular, a normalization constraint is applied on the latent variables,
\(\|\latent\|_2^2 = 1\) that restricts them to the surface of a unit sphere, encouraging comparability across samples and datasets while preventing degenerate solutions.

\subsection{Model architecture}
\label{sec:model-architecture}
The latent embedding model follows an encoder--decoder architecture, illustrated in \cref{fig:model-architecture}. The encoder maps input features to node-level latent variables, used to compute scene- and dataset-level embeddings. Two decoder heads are employed: one reconstructs input features, and the other predicts future states. The main building blocks of both networks are \glspl{GGRU}~\cite{westny2023mtp}.

\subsubsection{Graph-gated recurrent unit}
\label{sec:ggru}
The \gls{GGRU} enables spatio-temporal interactions to be captured by replacing the linear mappings in the \gls{GRU}~\cite{cho2014properties} with \gls{GNN} components. %
The original \gls{GGRU} design is simplified by using a minimal mechanism based on the minGRU~\cite{feng2024were}, reducing model complexity with minor performance trade-offs.

At each time step \(k\), a message-passing operation incorporates information from node \(i\) and its neighborhood \(\neigh{i}\), producing intermediate vectors:
\begin{equation}
    \left[\bm{v}_{k}^{(i)}, \bm{w}_{k}^{(i)} \right] = \text{GNN}\left(\bm{x}_k^{(i)}, \{\bm{x}_k^{(j)}\}_{j \in \neigh{i}}\right),
\end{equation}
where GNN$(\cdot)$ is the graph neural network operator from~\cite{morris2019weisfeiler}.
These are then used to update the hidden state \(\bm{h}_k^{(i)}\) using the following gating mechanism:
\begin{subequations}
\begin{align}
    \bm{z}_k^{(i)} &= \sigmoid(\bm{v}_{k}^{(i)} + \bm{b}_z), \\
    \tilde{\bm{h}}_k^{(i)} &= \tanh(\bm{w}_{k}^{(i)} + \bm{b}_{h}), \\
    \bm{h}_k^{(i)} &= (1 - \bm{z}_k^{(i)}) \odot \bm{h}_{k-1}^{(i)} + \bm{z}_k^{(i)} \odot \tilde{\bm{h}}_k^{(i)},
\end{align}
\end{subequations}
where \(\bm{b}_z\) and \(\bm{b}_h\) are learnable bias terms, \(\odot\) denotes the Hadamard product, and \(\sigmoid\) is the sigmoid function.

\label{sec:loss-function}
\subsection{Training objective}
The model is trained to minimize a combined objective of reconstruction and prediction losses. Node-level latent variables $\latent_i$ are inferred from input features $\ifeats_i$ using an encoder module $\encoder$. To ensure the representation captures structural data properties and relevant forecasting features, the total loss $\mathcal{L}$ is defined as
\begin{equation}
    \mathcal{L} = \|\recHead(\latentmat) - \ifeats\|^2 + \|\predHead(\latentmat) - \ofeats\|^2
\end{equation}
where $\recHead$ and $\predHead$ denote the reconstruction and prediction heads, respectively. Note that the decoder heads are solely used during training to shape the latent space and are discarded during inference. 

\subsection{Aggregating node- and scene-level representations}
Let $\mathbf{z}_{s,i}\in\mathbb{R}^{L}$ denote the representation of agent $i$ in scene $s$, where scene $s$ contains $n_s$ agents. The scene-level mean and covariance are computed as
\begin{equation}
    \boldsymbol{\mu}_s
    =
    \frac{1}{n_s}
    \sum_{i=1}^{n_s}\mathbf{z}_{s,i}, 
    \qquad
    \boldsymbol{\Sigma}_s
    =
    \frac{1}{n_s-1}
    \sum_{i=1}^{n_s}
    \left(\mathbf{z}_{s,i}-\boldsymbol{\mu}_s\right)
    \left(\mathbf{z}_{s,i}-\boldsymbol{\mu}_s\right)^{\mathsf T}.
\end{equation}

The aggregated dataset-level mean and covariance for a dataset $\mathcal{D}$ consisting
of $S$ scenes are defined as
\begin{equation}
    \label{eq:total_covariance}
    \bar{\boldsymbol{\mu}}_\mathcal{D}
    =
    \frac{1}{S}
    \sum_{s=1}^{S}\boldsymbol{\mu}_s,
    \qquad
    \boldsymbol{\Sigma}_{\mathcal{D}}
    =
    \underbrace{
    \frac{1}{S}
    \sum_{s=1}^{S}
    \left(\boldsymbol{\mu}_s-\bar{\boldsymbol{\mu}}_{\mathcal{D}}\right)
    \left(\boldsymbol{\mu}_s-\bar{\boldsymbol{\mu}}_{\mathcal{D}}\right)^{\mathsf T}
    }_{\text{Between-scene cov.}}
    +
    \underbrace{
    \frac{1}{S}
    \sum_{s=1}^{S}\boldsymbol{\Sigma}_s
    }_{\text{Within-scene cov.}}.
\end{equation}
The first covariance term captures variation between scene-level means, whereas the second captures the average variation among
agents within each scene.
Together, these terms provide a characterization of the dataset's latent structure and enable direct comparisons between datasets using probabilistic measures.

However, the empirical covariance may be ill-conditioned or singular, particularly in high-dimensional latent spaces with limited data. To improve numerical stability, a low-rank approximation with additive jitter regularization is used. Let
$\boldsymbol{\Sigma}_{\mathcal{D}} = \mathbf{U}\Lambda \mathbf{U}^\transpose$
denote the eigendecomposition of the covariance matrix, where the columns of $\mathbf{U}$ are the eigenvectors and $\Lambda$ is a diagonal matrix containing the corresponding eigenvalues in descending order. The regularized covariance is defined as
\begin{equation}
    \widetilde{\boldsymbol{\Sigma}}_{\mathcal{D}}
    =
    \mathbf{U}_r
    \operatorname{diag}
    \left(
        \lambda_1,\ldots,\lambda_r
    \right)
    \mathbf{U}_r^\transpose
    +
    \varepsilon \mathbf{I}, \qquad 
    \varepsilon
    =
    \alpha
    \frac{\operatorname{tr}(\boldsymbol{\Sigma}_{\mathcal{D}})}{L},
\end{equation}
where $\mathbf{U}_r$ contains the eigenvectors associated with the $r$ largest eigenvalues. The jitter parameter $\varepsilon$ is scaled according to the average marginal variance,
where $\alpha$ is a hyperparameter set to $3 \times 10^{-3}$ and $r$ is set to $16$, both values chosen based on empirical results (see the supplementary material for more details).

%% file: sec/results.tex
\section{Evaluation and results}
\label{sec:results}
This section evaluates the proposed framework. The datasets are introduced in \cref{sec:datasets}, followed by implementation details in \cref{sec:implementation-details}. The embedding model is assessed in \cref{sec:embedding-evaluation} through complementary quantitative and qualitative latent space analyses. \Cref{sec:transferability-evaluation} investigates how well latent distances predict cross-dataset generalization under zero-shot and fine-tuning settings. Finally, an ablation on the latent dimension $L$ analyzes its influence on transferability prediction accuracy.

\subsection{Datasets}
\label{sec:datasets}
The investigations are conducted on a diverse set of trajectory-prediction datasets spanning pedestrian, vehicle, and mixed-traffic domains, covering both map-based and map-free settings, and collected using static cameras, drones, or instrumented vehicles. An overview is provided in \cref{tab:datasets_summary}, where datasets are grouped by collection method and listed chronologically by release year.

All datasets are converted to a unified format using the \texttt{\small Dronalize} toolbox~\cite{westny2025toward}, including resampling to 10~Hz and adopting a common feature representation. Official train/validation/test splits are used when available; otherwise, the splitting protocol in \texttt{\small Dronalize} is applied.
For cross-dataset evaluation, all models are evaluated using a common 3~s prediction horizon, determined by the shortest fixed horizon among datasets with official test splits (\eg, View-of-Delft). During training, however, models predict up to a maximum horizon of 8~s, with supervision provided where future trajectories are available.

Following common practice~\cite{alahi2016social,gupta2018social}, the ETH and UCY datasets are organized into the five standard evaluation subsets: \emph{eth}, \emph{hotel}, \emph{univ}, \emph{zara1}, and \emph{zara2}.
Each subset is treated as a separate dataset throughout the experiments. Accordingly, the evaluation comprises 24 datasets in total.

\input{tab/dataset_table.tex}

\subsection{Implementation details}
\label{sec:implementation-details}
The embedding model was implemented in PyTorch~\cite{paszke2019pytorch} and PyTorch Geometric~\cite{fey2019pyg}. It was trained on a single NVIDIA A100 GPU using the AdamW optimizer~\cite{loshchilov2018decoupled} with a batch size of $512$.
To mitigate the imbalance in dataset sizes when jointly training on all datasets, mini-batches were constructed using weighted sampling, with each sample $j \in \mathcal{D}_i$ assigned the weight
\begin{equation}
    w_j \propto \frac{1}{|\mathcal{D}_i|^\alpha},
\end{equation}
where $\mathcal{D}_i$ denotes the $i$-th dataset. Here, $\alpha=0$ corresponds to sampling proportional to dataset size and $\alpha=1$ to uniform sampling across datasets.
A value of $\alpha=0.5$ is used, reducing the dominance of larger datasets while preserving a moderate bias toward them.
Training was conducted for $100$ epochs, with an initial learning rate of $10^{-3}$, which was decayed to $10^{-5}$
using a cosine annealing schedule~\cite{loshchilov2017sgdr}. Teacher forcing was used for both heads during the first 25
epochs, with the probability annealed linearly from 1 to 0.
The final models used in all the experiments were selected based on minimum ADE performance on the validation set.

\subsection{Embedding evaluation}
\label{sec:embedding-evaluation}
In the latent space, each dataset is modeled as a multivariate Gaussian distribution defined by its dataset-level mean
and covariance. Let $\mathcal{I}=\{1,\dots,K\}$ index the datasets $\dataset_1,\dots,\dataset_K$. For
$i,j\in\mathcal{I}$, the directed dissimilarity of $\dataset_i$ from $\dataset_j$ is defined as the \gls{KL}
divergence between their corresponding Gaussian distributions:
\begin{equation}
    \label{eq:kl}
    \kl(\dataset_i \| \dataset_j)
    =
    \kl\!\left(
        \mathcal{N}\!\left(
            \bar{\boldsymbol{\mu}}_{\dataset_i},
            \widetilde{\boldsymbol{\Sigma}}_{\dataset_i}
        \right)
        \,\middle\|\,
        \mathcal{N}\!\left(
            \bar{\boldsymbol{\mu}}_{\dataset_j},
            \widetilde{\boldsymbol{\Sigma}}_{\dataset_j}
        \right)
    \right),
    \qquad i,j\in\mathcal{I},
\end{equation}
where $\kl(\dataset_i \| \dataset_j)$ measures how well the distribution of dataset $\dataset_j$ approximates that of
dataset $\dataset_i$. Since the \gls{KL} divergence is asymmetric, the resulting dissimilarity captures
directional differences in dataset coverage and variability.

\begin{figure*}[t]
    \centering
    \includegraphics[width=1\textwidth]{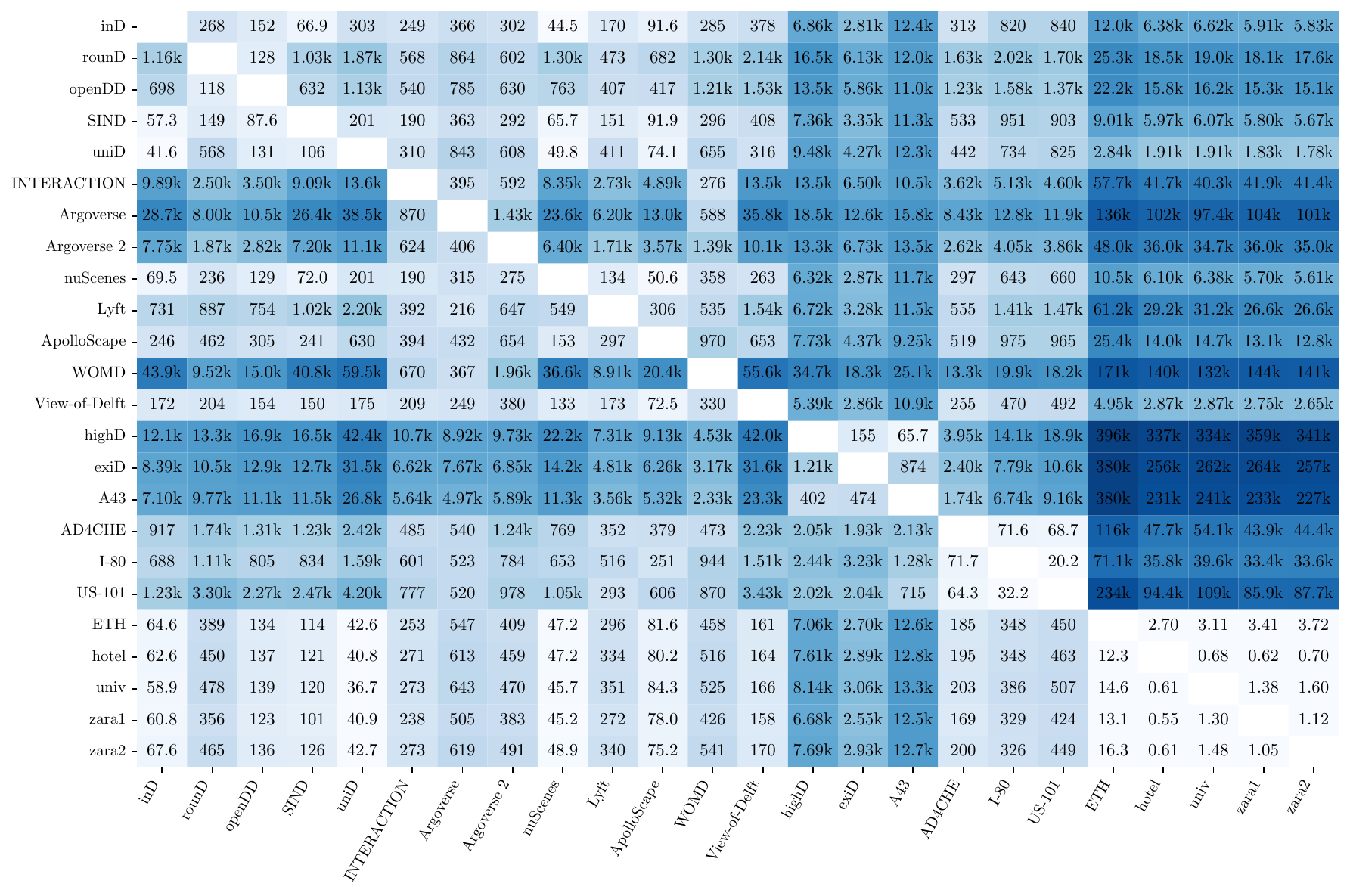}
    \vspace{-0.65cm}
    \caption{KL divergence $\kl(\text{row}\|\text{col})$ between dataset embedding distributions. Entries indicate how well column datasets approximate row datasets. Lower values denote closer alignment, typically correlating with stronger zero-shot transfer from column training datasets to row evaluation datasets. Matrix asymmetry reflects directional differences in dataset coverage and variability.}
    \label{fig:kl-div-fig}
\end{figure*}

\begin{figure}[t]
    \centering
    \begin{subfigure}[b]{0.42\textwidth}
        \centering
        \includegraphics[width=\linewidth]{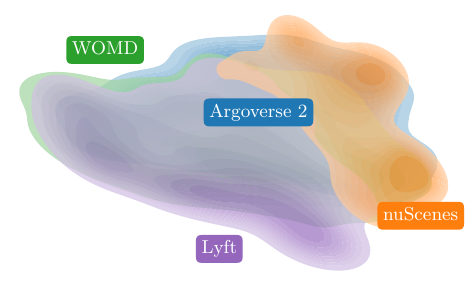}
        \caption{Urban datasets}
        \label{fig:urban-pca-fig}
    \end{subfigure}
    \hfill
    \begin{subfigure}[b]{0.42\textwidth}
        \centering
        \includegraphics[width=\linewidth]{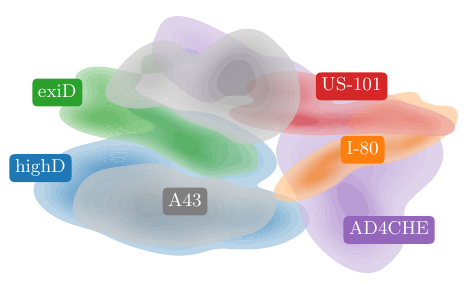}
        \caption{Highway datasets}
        \label{fig:hw-pca-fig}
    \end{subfigure}
    
    \caption{Two-dimensional t-SNE visualization of the learned scenario embeddings for (a) a subset of the urban datasets (acquired with instrumented vehicles) and (b) highway datasets, with contour lines indicating density.}
    \label{fig:combined-tsne}
\end{figure}

\Cref{fig:kl-div-fig} presents a heatmap of the \gls{KL} divergence between embedding distributions for each dataset
pair, where lower values indicate greater latent similarity. Several interesting patterns emerge. For example, nuScenes
appears \emph{easier} to approximate using other instrumented-vehicle datasets such as Lyft L5, WOMD, Argoverse, and
Argoverse~2 than vice versa. For instance,
\begin{align}
	D_{\text{KL}}(\text{nuScenes} \| \text{WOMD}) = 358, \quad
	D_{\text{KL}}(\text{WOMD} \| \text{nuScenes}) = 36602,
\end{align}
indicating that these datasets contain information that transfers to nuScenes more effectively than nuScenes transfers
to them. This asymmetry aligns with the empirical findings in both~\cite{feng2024unitraj} and
\cite{gilles2022uncertainty}, where models pretrained on WOMD, Argoverse, or Argoverse 2 and evaluated on nuScenes
achieve notably stronger performance than the reverse.
The t-SNE~\cite{maaten2008visualizing} projection of scene latents in \cref{fig:urban-pca-fig} supports this:
nuScenes occupies a distinct subregion within the other datasets' broader clusters. This indicates that the asymmetric
relationship observed in the high-dimensional latent space is also reflected in a low-dimensional projection.

Highway datasets are more similar to each other than to urban datasets, though asymmetrically.
The t-SNE projection in \cref{fig:hw-pca-fig} clearly shows overlap among highD, exiD, and A43, all collected in Germany, likely reflecting
regional driving patterns.
Furthermore, AD4CHE, I-80, and US-101 cluster together, suggesting the latent space encodes shared behavioral or traffic-flow characteristics beyond geographic proximity.

Another important observation is the large divergence between the pedestrian-focused datasets and the rest. The results
indicate that the pedestrian datasets are substantially different from the others, suggesting low potential for direct
transferability. Partial exceptions are uniD and View-of-Delft, which are heavily pedestrian-oriented and appear
to be closer in the embedding space.

Most interestingly, the embedding space also uncovers relationships between datasets that do not share presumed
commonalities, such as acquisition method or agent type. For instance, several drone-acquired datasets, including inD,
openDD, SIND, INTERACTION, exhibit low \gls{KL} divergence with instrumented-vehicle datasets such as nuScenes,
Argoverse~2, and WOMD. This suggests that cross-domain transfer is possible even between heterogeneous datasets.
This is significant, as trajectory-prediction research often suffers from scarce, domain-specific
data~\cite{wang2025deployable}, indicating untapped potential in leveraging datasets that are comparatively underused in
the motion prediction literature.

\subsection{Transferability evaluation}
\label{sec:transferability-evaluation}

To evaluate the interpretation of \emph{closeness} in the latent space as a measure of transferability, a series of
experiments is conducted in which a trajectory-prediction model is trained on one dataset and evaluated on another.
Motivated by its state-of-the-art performance and its architectural \textbf{dissimilarity} to the proposed embedding
model, QCNet~\cite{zhou2023query} is adopted as the baseline predictor, following the original training objectives and
inference procedure. To ensure comparability across datasets, a lightweight input transformation layer is introduced,
and model capacity is reduced to maintain computational feasibility across all datasets (see the supplementary material
for more details). Model performance is reported using the minimum average positional displacement error for the best of
$K = 6$ predicted trajectories (minADE$_6$) over a 3-second horizon.

\subsubsection{Zero-shot transferability}

To evaluate cross-dataset generalization, a model is trained on one dataset and is evaluated, without any adaptation, on
all remaining datasets. With $24$ datasets, this yields $24 \times 23 = 552$ transfer pairs.

\begin{figure}[t]
    \centering
    \includegraphics[width=0.5\linewidth]{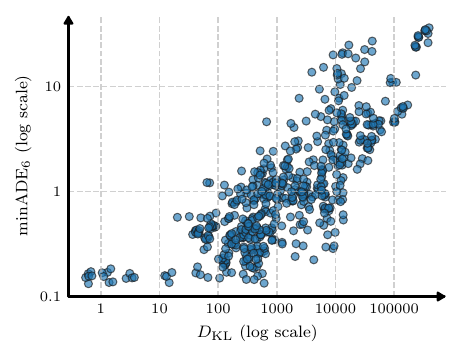}
    \caption{KL divergence versus zero-shot minADE$_6$ transfer performance across all $552$ dataset pairs. Larger divergence correlates with worse transferability.}
    \label{fig:correlation-kl-zeroshot-fig}
\end{figure}

Using the divergence introduced in \eqref{eq:kl}, it is investigated how well this measure
predicts zero-shot transfer performance. To this end, minADE$_6$ is examined against
$\kl(\dataset_e \| \dataset_t)$, where $\dataset_e$ and $\dataset_t$ denote evaluation and training datasets,
respectively. As shown in \cref{fig:correlation-kl-zeroshot-fig}, pairs with larger divergence exhibit consistently
higher minADE$_6$, indicating poorer transfer.

A Spearman's rank correlation analysis confirms a strong positive association: increasing divergence aligns with reduced
zero-shot performance. Overall, the KL divergence achieves a rank correlation of $\rho = 0.811$, with a $95\%$
confidence interval of $(0.782,\,0.840)$.

\subsubsection{Transferability under fine-tuning}
Beyond zero-shot performance, it is important to understand how datasets support transfer when used for fine-tuning
toward a target domain, as this has direct practical utility.
This is studied from two complementary perspectives: whether the latent divergence can guide pretraining source
selection, and whether it predicts how much source knowledge is lost during adaptation.

\begin{figure}[t]
    \centering
    \begin{minipage}[t]{0.48\textwidth}
        \centering
        \includegraphics[width=0.82\linewidth]{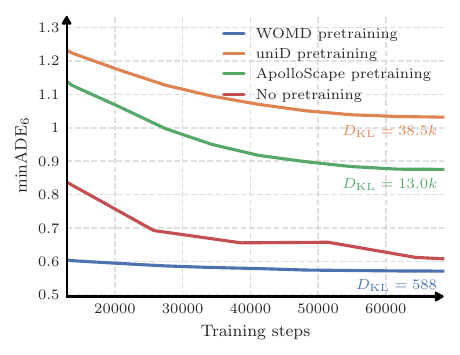}
        \caption{Argoverse fine-tuning from different sources. Sources with lower divergence scores yield better adaptation.}
        \label{fig:finetuning}
    \end{minipage}
    \hfill
    \begin{minipage}[t]{0.48\textwidth}
        \centering
        \includegraphics[width=0.82\linewidth]{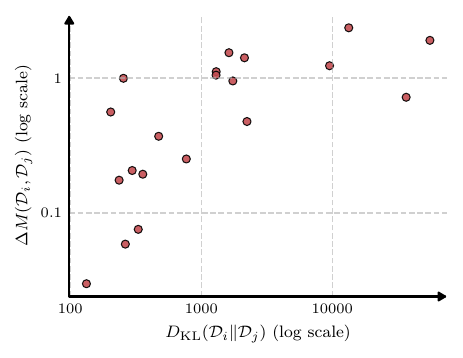}
        \caption{KL divergence versus performance degradation after fine-tuning. Higher divergence generally corresponds to greater forgetting of the source dataset.}
        \label{fig:forgetting-kl-finetune-fig}
    \end{minipage}
\end{figure}

\paragraph{Pretraining source selection.}
As a practical case study, Argoverse is used as the target dataset. Models pretrained on a set of candidate source
datasets are subsequently fine-tuned on Argoverse, and target performance is recorded after adaptation. As shown in
\cref{fig:finetuning}, sources with lower \gls{KL} divergence to Argoverse generally yield stronger
post-fine-tuning performance on the target domain, indicating that the latent metric can serve as a practical guide
for pretraining source selection, even before any task-specific training is conducted.

\paragraph{Catastrophic forgetting.}
A complementary concern when fine-tuning is catastrophic forgetting~\cite{goodfellow2013empirical}: the
tendency for adaptation to a new domain to overwrite previously acquired source knowledge. To examine this, models
pretrained on a source dataset $\dataset_i$ are fine-tuned on a target dataset $\dataset_j$, and performance is
subsequently evaluated on $\dataset_i$. Let $m_{\text{org}}^{\mathcal{D}_i}$ and $m_{\text{ft}}^{\mathcal{D}_j}$ denote the original and
fine-tuned models, respectively. The performance change is defined as:
\begin{equation}
\Delta M(\dataset_i, \dataset_j) = M(m_{\text{ft}}^{\mathcal{D}_j}, \dataset_i) - M(m_{\text{org}}^{\mathcal{D}_i}, \dataset_i),
\end{equation}
where $M(\cdot, \cdot)$ denotes minADE$_6$ evaluated on a specific dataset.
Lower $\Delta M$ indicates less forgetting, and it is expected to correlate with $D_{\text{KL}}(\dataset_i \| \dataset_j)$.

Experiments across five datasets, AD4CHE, nuScenes, rounD, View-of-Delft, and WOMD, produce $20$ transfer pairs.
\Cref{fig:forgetting-kl-finetune-fig} indicates a positive trend: larger divergence between datasets generally
coincides with stronger forgetting. Spearman's rank correlation is $\rho=0.729$, with a $95$\%
confidence interval of $(0.403,\,0.913)$, confirming a strong positive association.
Taken together, the two results show that the latent divergence is predictive of fine-tuning behavior from both directions: it anticipates adaptation
quality before training, and captures the degree of source knowledge loss afterward.

\subsubsection{Motivation for the Gaussian assumption}

To assess the Gaussian modeling assumption, \cref{tab:metric_dimension_comparison} compares the predictive performance of the \gls{KL}
divergence with alternative similarity measures, including the non-parametric \gls{MMD}~\cite{gretton2012kernel}
computed directly on scene-level embeddings. The \gls{KL} divergence consistently achieves the highest rank
correlation across all evaluated latent dimensionalities in both zero-shot and fine-tuning tasks, empirically justifying
the Gaussian approximation.

\input{tab/alt_metrics}

\subsubsection{Ablation on latent dimension size}
Finally, an ablation study examines how the latent dimensionality $L$ influences the predictive quality of the learned
embeddings. Models with $L \in \{16, 32, 64, 128\}$ are evaluated on both zero-shot transfer and post-fine-tuning
degradation tasks. As shown in \cref{tab:metric_dimension_comparison}, the two tasks exhibit opposing trends. For zero-shot
transfer, the correlation decreases from $\rho=0.811$ at $L=32$ to $\rho=0.736$ at $L=128$, indicating that the more
compact representation better preserves the ordering of cross-dataset transfer performance. In contrast, the correlation
for post-fine-tuning degradation increases from $\rho=0.729$ to $\rho=0.874$, suggesting that larger latent spaces may
better capture differences associated with adaptation and forgetting. However, the wide and overlapping confidence
intervals, together with the smaller number of fine-tuning pairs, warrant a more cautious interpretation of this trend.

\subsection{Investigation into latent space quality}
To evaluate whether latent dataset embeddings capture complex discrepancies beyond basic statistical heuristics, transfer performance $T_{i,j}$ from source dataset $\dataset_i$ to target $\dataset_j$ is modeled using the KL divergence and a set of candidate explicit heuristic metrics $\gamma^{(\ell)}_{i,j}$
\begin{equation}
    \label{eq:linear_model}
    T_{i,j} \approx a\,D_{\mathrm{KL}}(\dataset_j \| \dataset_i) + \sum_{\ell \in \mathcal{S}} b_{\ell}\,\gamma^{(\ell)}_{i,j} + c.
\end{equation}
Here, $\mathcal{S}$ encompasses five dataset differences, with definitions and correction ratios detailed in \cref{tab:correction_terms}. To isolate the most informative predictors, this model is optimized using Lasso regularization, applying an $L_1$ penalty to the coefficients $a$ and $b_{\ell}$. Formally, the parameters $\theta = \{a, \{b_{\ell}\}_{\ell \in \mathcal{S}}, c\}$ are estimated by minimizing
\begin{equation}
    \min_{\theta}\; \sum_{i,j} \Bigl( T_{i,j} - a\,D_{\mathrm{KL}}(\dataset_j \| \dataset_i) - \sum_{\ell \in \mathcal{S}} b_{\ell}\,\gamma^{(\ell)}_{i,j} - c \Bigr)^{2} + \lambda \Bigl( |a| + \sum_{\ell \in \mathcal{S}} |b_{\ell}| \Bigr).
\end{equation}
This formulation enforces sparsity. Coefficients driven to zero correspond to predictors that do not contribute substantially to explaining transfer performance. 

\input{tab/dataset_characteristic.tex}

\begin{figure*}[t]
    \centering
    \begin{subfigure}{0.43\textwidth}
        \centering
        \includegraphics[width=\linewidth]{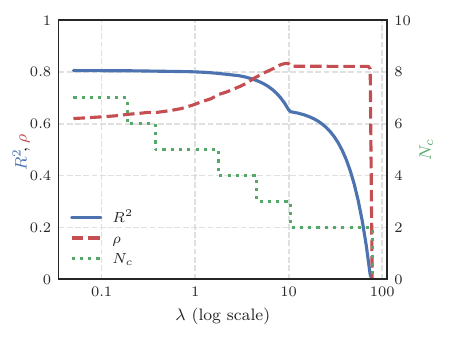}
        \caption{}
        \label{fig:first}
    \end{subfigure}
    \hfill
    \begin{subfigure}{0.53\textwidth}
        \centering
        \includegraphics[width=\linewidth]{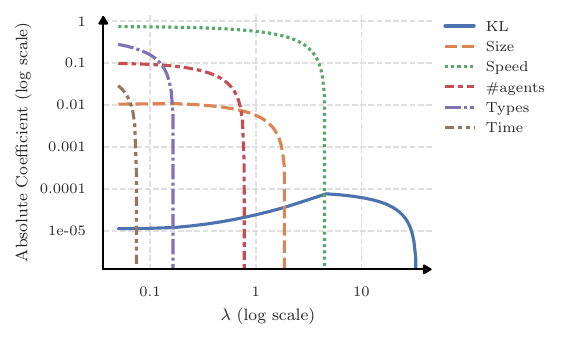}
        \caption{}
        \label{fig:second}
    \end{subfigure}
    \vspace{-0.2cm}
    \caption{
    (a) Evolution of coefficient of determination $R^2$, Spearman's $\rho$, and the number of selected correction terms $N_c$ against the regularization parameter $\lambda$.  
    (b) Coefficient paths for all terms in~\eqref{eq:linear_model}.}
    \label{fig:combined}
\end{figure*}

The regression results in \cref{fig:combined} demonstrate the representational capacity of the learned latent space. In
\cref{fig:second}, the KL divergence coefficient remains nonzero for much higher values of the regularization
strength $\lambda$ compared to the other coefficients, indicating that the learned representation captures these
underlying patterns.

The exception is average agent speed, as its coefficient remains nonzero under stronger regularization. Nevertheless,
using the speed factor by itself achieves a lower rank correlation, $\rho=0.624$, with a $95\%$ confidence interval of
$(0.561,\,0.669)$, compared with $\rho=0.811$ for the KL divergence, with a $95\%$ confidence interval of
$(0.782,\,0.840)$. This further indicates that the latent representation captures discrepancies that are not adequately
described by the considered statistical heuristics.

%% file: tab/dataset_table.tex
\begin{table*}[ht]
    \centering
    \caption{
        Overview of the 24 trajectory-prediction datasets used in the experiments.
        Locations are condensed to country. Map info indicates whether HD maps
        (\cmark), no maps (\xmark), or partial map data are provided. The datasets
        balance prevalence in the literature with complementary characteristics,
        including geographic coverage and agent composition.
    }
    \label{tab:datasets_summary}
    \renewcommand{\arraystretch}{1.05}
    \setlength{\tabcolsep}{5pt}
    \scriptsize
    \resizebox{\textwidth}{!}{%
    \begin{tabular}{lclllcS[table-format=6.0]}
        \toprule
        \textbf{Name}
        & \textbf{Year}
        & \textbf{Location}
        & \textbf{Agents}
        & \textbf{Sensor}
        & \textbf{Map Info}
        & \textbf{Size} \\
        \midrule

        I-80~\cite{i80v2}
        & 2006 & USA & Vehicles & Camera & Lane lines & 23480 \\

        US-101~\cite{us101v2}
        & 2007 & USA & Vehicles & Camera & Lane lines & 18683 \\

        univ~\cite{lerner2007crowds}
        & 2007 & Cyprus & Pedestrians & Camera & \xmark & 4196 \\

        zara1~\cite{lerner2007crowds}
        & 2007 & Cyprus & Pedestrians & Camera & \xmark & 4096 \\

        zara2~\cite{lerner2007crowds}
        & 2007 & Cyprus & Pedestrians & Camera & \xmark & 4103 \\

        eth~\cite{pellegrini2009you}
        & 2009 & Switzerland & Pedestrians & Camera & \xmark & 4013 \\

        hotel~\cite{pellegrini2009you}
        & 2009 & Switzerland & Pedestrians & Camera & \xmark & 4040 \\

        \midrule

        ApolloScape~\cite{ma2019trafficpredict}
        & 2019 & China & Mixed & Vehicle & \xmark & 49361 \\

        Argoverse~\cite{chang2019argoverse}
        & 2019 & USA & Mixed & Vehicle & \cmark & 323557 \\

        nuScenes~\cite{caesar2020nuscenes}
        & 2020 & USA, Singapore & Mixed & Vehicle & \cmark & 195103 \\

        Lyft Level 5~\cite{houston2021one}
        & 2021 & USA & Mixed & Vehicle & \cmark & 292329 \\

        WOMD~\cite{ettinger2021large}
        & 2021 & USA & Mixed & Vehicle & \cmark & 576012 \\

        Argoverse 2~\cite{argoverse2dataset}
        & 2023 & USA & Mixed & Vehicle & \cmark & 249880 \\

        View-of-Delft~\cite{boekema2024vodp}
        & 2024 & Netherlands & Mixed & Vehicle & \cmark & 10486 \\

        \midrule

        highD~\cite{highDdataset}
        & 2018 & Germany & Vehicles & Drone & Lane lines & 129786 \\

        INTERACTION~\cite{zhan2019interaction}
        & 2019
        & \makecell[l]{USA, China,\\Germany, Bulgaria}
        & Mixed & Drone & \cmark & 62022 \\

        inD~\cite{inDdataset}
        & 2020 & Germany & Mixed & Drone & \cmark & 155609 \\

        rounD~\cite{rounDdataset}
        & 2020 & Germany & Mixed & Drone & \cmark & 111973 \\

        openDD~\cite{breuer2020opendd}
        & 2020 & Germany & Mixed & Drone & \cmark & 370073 \\

        exiD~\cite{exiDdataset}
        & 2022 & Germany & Vehicles & Drone & \cmark & 311309 \\

        SIND~\cite{sinDdataset}
        & 2022 & China & Mixed & Drone & \cmark & 260484 \\

        AD4CHE~\cite{zhang2023ad4che}
        & 2023 & China & Vehicles & Drone & Images & 41333 \\

        uniD~\cite{uniDdataset}
        & 2024 & Germany & Mixed & Drone & \cmark & 189023 \\

        A43~\cite{a43dataset}
        & 2024 & Germany & Vehicles & Drone & Lane lines & 44565 \\

        \bottomrule
    \end{tabular}%
    }
\end{table*}

%% file: tab/alt_metrics.tex
\newcommand{\klcell}[3]{%
  \shortstack{%
    $\mathbf{#1}$\\[-1pt]
    \makebox[0pt][c]{\tiny $(#2,#3)$}%
  }%
}

\begin{table}[t]
  \centering
  \caption{
    Spearman's rank correlation coefficient $\rho$ for different dataset
    dissimilarity measures and latent dimensions. \textbf{Bold} denotes the best
    measure for each task--dimension combination, where higher is better. For KL divergence,
    $95\%$ confidence intervals are shown below the corresponding estimates.
  }
  \label{tab:metric_dimension_comparison}
  \setlength{\tabcolsep}{5.9pt}
  \scriptsize
  \renewcommand{\arraystretch}{1.15}

  \begin{tabular}{
      l
      cccc
      !{\color{lightgray}\vline}
      cccc
    }
    \toprule
    & \multicolumn{4}{c}{\textbf{Zero-shot transfer}}
    & \multicolumn{4}{c}{\textbf{Post-fine-tuning degradation}} \\
    \cmidrule(lr){2-5}
    \cmidrule(lr){6-9}
    \textbf{Measure}
      & $L=16$ & $L=32$ & $L=64$ & $L=128$
      & $L=16$ & $L=32$ & $L=64$ & $L=128$ \\
    \midrule

    L1
      & $0.475$ & $0.468$ & $0.470$ & $0.425$
      & $0.136$ & $0.254$ & $0.226$ & $0.145$ \\

    Wasserstein
      & $0.489$ & $0.482$ & $0.482$ & $0.448$
      & $0.269$ & $0.296$ & $0.335$ & $0.257$ \\

    MMD
      & $0.410$ & $0.412$ & $0.410$ & $0.362$
      & $0.136$ & $0.245$ & $0.193$ & $0.185$ \\

    \midrule

  KL
    & \klcell{0.781}{.750}{.810}
    & \klcell{0.811}{.782}{.840}
    & \klcell{0.746}{.711}{.783}
    & \klcell{0.736}{.681}{.760}
    & \klcell{0.504}{.045}{.792}
    & \klcell{0.729}{.403}{.913}
    & \klcell{0.794}{.578}{.922}
    & \klcell{0.874}{.639}{.970} \\

    \bottomrule
  \end{tabular}
\end{table}

%% file: tab/dataset_characteristic.tex
\begin{table}[t]
    \centering
    \caption{Definitions of dataset characteristics and derived correction terms $\gamma^{(\ell)}_{i,j}$ for transfer from
    source $\dataset_i$ to target $\dataset_j$. Scalar corrections are computed as the ratio of source to target values,
    whereas the agent class distribution $\mathbf{p} \in \mathbb{R}^2$ use $L_1$ distance.}
    \label{tab:correction_terms}
    \scriptsize
    \renewcommand{\arraystretch}{1.05}
    \setlength{\tabcolsep}{4.2pt}
    \begin{tabular}{l l c c}
        \toprule
        \textbf{Term} & \textbf{Description} & \textbf{Symbol} & \textbf{Calculation of $\gamma^{(\ell)}_{i,j}$} \\
        \midrule
        Size
        & Total number of samples in the dataset
        & $N$
        & $N_i \,/\, N_j$ \\

        Speed
        & Mean agent speed across samples and time steps
        & $\bar{v}$
        & $\bar{v}_i \,/\, \bar{v}_j$ \\

        $\#$agents
        & Mean number of agents present in each sample
        & $\bar{n}$
        & $\bar{n}_i \,/\, \bar{n}_j$ \\

        Types
        & Mean empirical distribution over agent classes
        & $\bar{\mathbf{p}}$
        & $\|\bar{\mathbf{p}}_i - \bar{\mathbf{p}}_j\|_1$ \\

        Time
        & Mean duration of the observed input sequence
        & $\bar{t}$
        & $\bar{t}_i \,/\, \bar{t}_j$ \\
        \bottomrule
    \end{tabular}
\end{table}

%% file: sec/conclusion.tex
\section{Conclusion}
A latent embedding framework that systematically characterizes trajectory datasets and quantifies their transferability
was introduced. The framework reveals structured relationships between datasets and offers guidance on selecting
effective pretraining sources. Empirical results across 24 diverse datasets show that distances in the latent space
strongly correlate with both zero-shot generalization and fine-tuning outcomes, demonstrating that transferability can
be predicted rather than discovered through exhaustive experimentation. The analysis further identifies datasets that
are less commonly used in prior work but still hold potential to improve performance on widely used motion-prediction
benchmarks. These findings offer practical guidance for dataset selection and model development, contributing toward
more robust and generalizable motion prediction systems.

%% file: sec/suppl.tex
\clearpage
\appendix

\section{Latent embedding model}
This section expands on the latent embedding model and provides additional details on its architecture and hyperparameters.

\subsection{Network configuration}

As shown in \cref{fig:model-architecture}, the latent embedding model comprises five main components:
\begin{enumerate}
    \item \gls{GGRU} encoder for agent features and interactions
    \item GNN-based map encoder for processing the lane graph
    \item Feed-forward module that fuses agent and map features into the latent embedding~$\latentmat$
    \item \gls{GGRU} decoder head for reconstructing the input
    \item \gls{GGRU} decoder head for trajectory forecasting
\end{enumerate}
All modules use a hidden dimension of~$128$, and all graph operations rely on the GNN operator of~\cite{morris2019weisfeiler}.

Following a simple embedding layer that includes agent type information, the \gls{GGRU} encoder processes agent features sequentially for each instance, performing message passing between neighboring agents.
The map encoder operates in two stages. First, it performs message passing on the lane-graph to compute node-level map embeddings.
The raw map inputs consist of node positions and dataset-defined lane-graph connectivity.
Edge features include the map-node type specified by the \texttt{\small Dronalize} toolbox~\cite{westny2025toward}; these are one-hot encoded, projected through a linear layer, and processed by a three-layer GNN.
Second, a directed three-layer GNN propagates information from map nodes to agents.
The resulting map features are then passed through a LayerNorm layer~\cite{ba2016layer}.

The outputs of the \gls{GGRU} encoder and the map encoder are concatenated and passed through a two-layer feed-forward network with SiLU activations~\cite{hendrycks2016gaussian} to produce the latent embedding~$\latentmat$, followed by normalization.
This embedding initializes the hidden state of the two \gls{GGRU} decoder heads, which autoregressively reconstruct the input and predict future trajectories.
At each step, the decoders take their own previous prediction, transformed by a two-layer feed-forward network with SiLU activations, as the next input.

\subsection{Graph construction}
Graphs are constructed on a per-sample basis using heuristic rules tailored to capture relevant spatial and semantic structure.
Inter-agent graphs are built by connecting each agent to others within a 150~m radius, limited to a maximum of 10 neighbors per node. All edges are undirected, with edge weights computed as the exponential of the negative relative distance.
Map-agent graphs are constructed from the lane graph, if available, which is initially represented as a set of polylines with distinctive types.
These polylines are discretized into graph nodes by sampling points at approximately 3~m intervals. Agents are then connected to nearby map nodes using a nearest-neighbor criterion, allowing up to 50 directed edges from map nodes to agents. For both graph types, edge dropout~\cite{rong2020drop}
 is applied during training with a probability of 0.25.

\section{Trajectory prediction model}

QCNet~\cite{zhou2023query} serves as the baseline predictor for the transferability experiments.
While the original implementation is largely followed, a few modifications were made to adapt it to this work.

\input{tab/qcnet_params.tex}

\subsection{Network configuration}
The only functional modification concerns the map encoder, which was adapted to match the input format defined by the \texttt{\small Dronalize} toolbox~\cite{westny2025toward}. In practice, this required adjusting the architecture to accommodate a different set of map-node types and edge definitions.
To reduce computational demands and training time across the extensive experimental suite, several hyperparameters of
the original QCNet model were scaled down. \Cref{tab:qcnet-hyperparams} summarizes the modified settings alongside the
original values. These adjustments lower the model size from roughly 7.66 million parameters to 1.35 million, yielding
substantial reductions in memory usage and training duration. %
Initial studies showed that these changes had only minor effects on performance and, where differences appeared, they followed the same trends observed with the full-size model, making the compact configuration sufficient for the purposes of the transferability analyses of this work.

\subsection{Training details}

Models are trained for up to $50$ epochs using AdamW~\cite{loshchilov2018decoupled}. Batch sizes and
learning rates are selected separately for each dataset (see \cref{tab:qcnet-training-config}). Early stopping is
applied based on the validation minADE$_6$, with a patience of $5$ epochs. Given the number of datasets and total runs,
learning-rate tuning is limited. The reported values represent stable, workable settings rather than the result of an
extensive hyperparameter search. As for the latent embedding model, the final QCNet models used in all experiments were
selected based on minADE$_6$ performance on the validation set.
\input{tab/qcnet_training.tex}

\section{Experiments}
This section presents additional experiments that complement the main results. These include extended analyses, alternative evaluation settings, and further qualitative examples that provide additional insight into the behavior of the proposed framework.

\subsection{Recommender system}

One objective of the latent embedding model is to provide a principled, data-driven mechanism for identifying suitable
source datasets for pre-training or adaptation. When a new dataset becomes available, its compatibility with existing
sources can be assessed by computing its embedding and comparing it with the existing embeddings.

\begin{figure*}[t]
	\centering
	\includegraphics[width=1\textwidth]{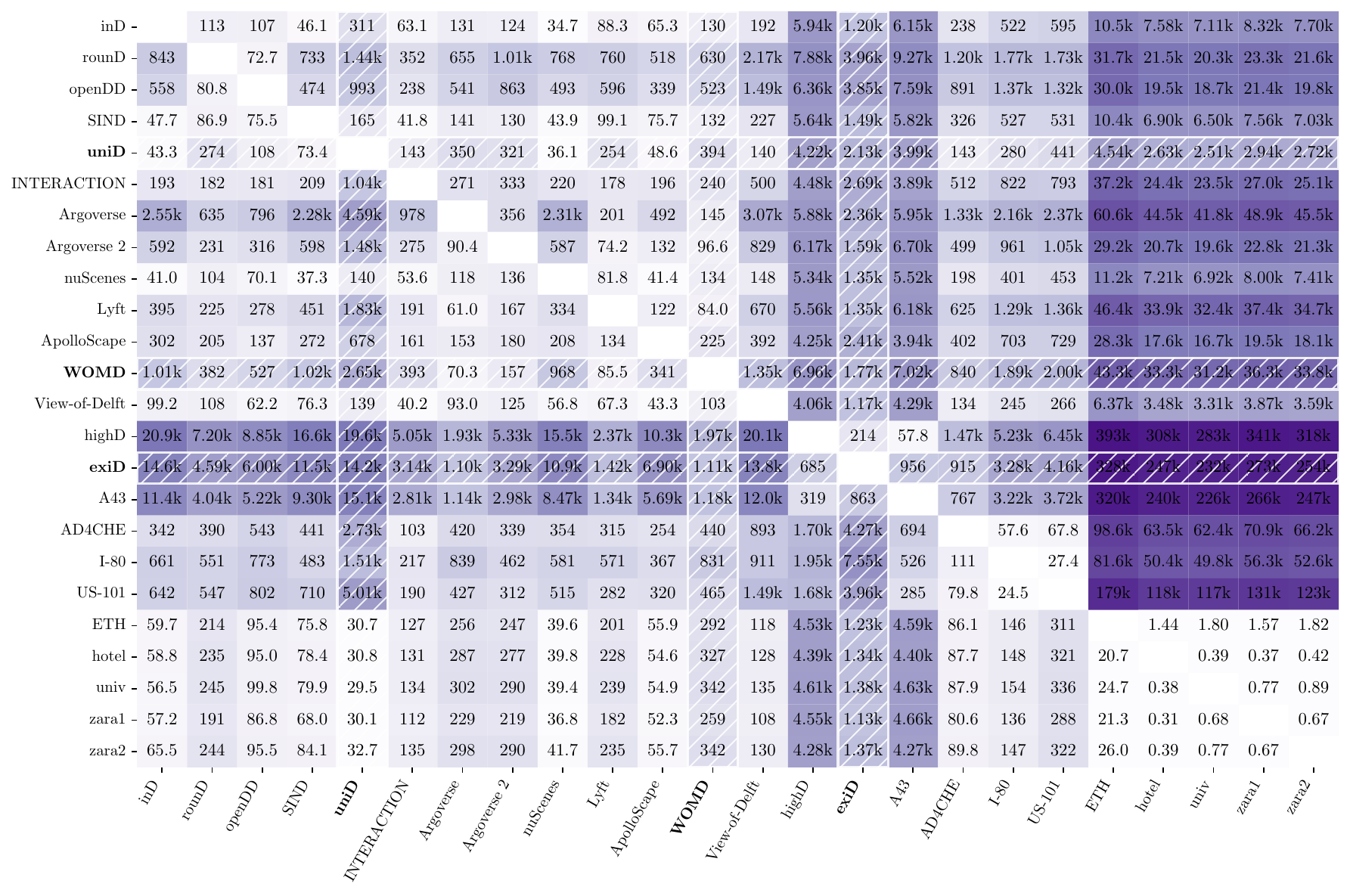}
  \vspace{-0.65cm}
	\caption{KL divergence $\kl(\text{row}\|\text{col})$ between latent embedding distributions of all
	dataset pairs when WOMD, exiD, and uniD are excluded from training. Each entry measures how well the column dataset
	can approximate the row dataset in latent space, where lower values indicate better approximation. Datasets excluded
    from training are highlighted.}
    \label{fig:kl-div-except-fig}
\end{figure*}

To illustrate this process, an unseen-data scenario is simulated by training a new embedding model after removing WOMD,
exiD, and uniD from the training set. The resulting \gls{KL} divergence heatmap from this model is shown in
\cref{fig:kl-div-except-fig}. Despite the exclusion, the model positions WOMD near Argoverse, uniD near inD, and exiD
near highD---relationships that closely match those in the full-training heatmap (\cref{fig:kl-div-fig}). Although the
similarity to the fully trained model may appear unsurprising, it is precisely this consistency that is meaningful: the embedding
space recovers the same structural relationships even without direct exposure to the omitted datasets.

This stability suggests that the learned embedding generalizes well to unseen data and that it can reliably support both
transferability assessment and source-dataset recommendation.

\subsection{Effect of latent dimensionality on correlation dynamics}

\begin{figure*}[t]
    \centering
    \includegraphics[width=0.7\textwidth]{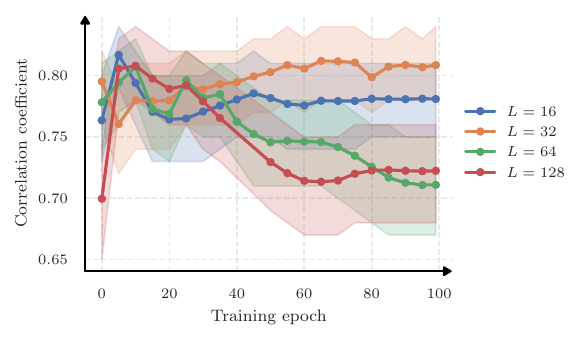}
    \vspace{-0.185cm}
    \caption{Evolution of Spearman's rank correlation coefficient $\rho$ over training epochs for models with latent
    dimensionalities $L \in \{16, 32, 64, 128\}$, illustrating how capacity influences the stability and growth of learned
    correlations.}
    \label{fig:latent_space_epoch_vs_rho_all}
\end{figure*}

Figure~\ref{fig:latent_space_epoch_vs_rho_all} illustrates how the latent dimensionality $L$ influences the evolution of
the learned correlations during training. The $L=32$ model achieves the highest and most stable Spearman's $\rho$, while
the $L=16$ model performs slightly worse. Increasing the latent dimension to $L=64$ or $L=128$ does not provide
additional benefit and results in lower final correlations. In particular, the $L=128$ model exhibits a downward trend,
which may indicate overfitting. These results suggest that increasing the latent capacity beyond $L=32$ may amplify noise
rather than improving the learned representation.

\subsection{Transferability results}

\begin{figure*}[t]
    \centering
    \includegraphics[width=0.99\textwidth]{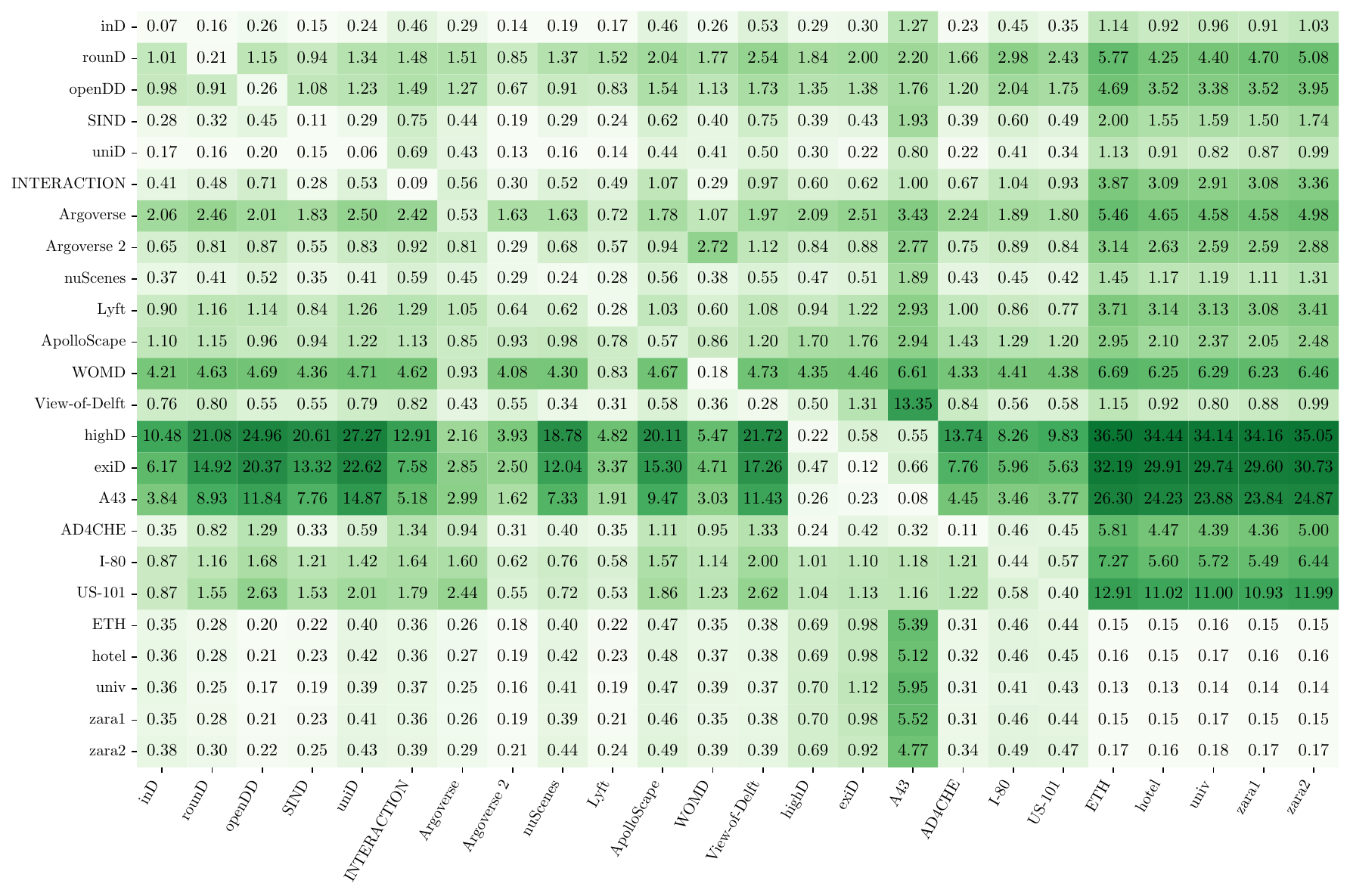}
    \vspace{-0.4cm}
    \caption{Cross-dataset zero-shot minADE$_6$ results for a 3~s prediction horizon. Columns indicate training datasets, and rows indicate evaluation datasets. Lower values reflect better transferability.}
    \label{fig:eval_matrix}
\end{figure*}

For completeness, the full zero-shot transferability results are provided in \cref{fig:eval_matrix}, showing
minADE$_6$ performance over a 3\,s prediction horizon. Several patterns emerge. First, transferring from urban sources
to highway targets is noticeably more difficult than the reverse.

Although transferring from urban sources to highway targets is challenging, highway prediction itself is inherently
easier once trained on in-domain data. This is reflected in the relatively low displacement errors observed for
highway-to-highway transfer (\eg, exiD, highD, A43). In contrast, urban-to-urban transfer proves substantially more
difficult, with datasets such as rounD and openDD showing markedly higher errors despite the within-domain setting.
This aligns with prior empirical findings~\cite{westny2023eval}, which similarly report that highway scenarios are
generally easier to predict.

One row that stands out is WOMD, which appears to be one of the most challenging target datasets. Only Argoverse and
Lyft achieve reasonably competitive zero-shot performance. On the other hand, WOMD is a strong source dataset, offering
solid transfer to many targets. Lyft and Argoverse~2 show similar behavior. As noted in the main text, these datasets
are also among the largest in the study, covering broad regions of the latent space and thus providing strong potential
for pre-training.

Finally, several datasets transfer surprisingly well to the pedestrian-centric ones. This should, however, be
interpreted with caution, as these are predominantly low-speed scenarios where best-of-$K$ metrics tend to overestimate
performance~\cite{scholler2020constant,mohamed2022social,weng2023joint}. Regardless, it is noteworthy that pre-training
on highway datasets can serve as an effective starting point for zero-shot or fine-tuned pedestrian prediction.

\subsection{Single-source selection for zero-shot transfer}

To assess the practical utility of the proposed \gls{KL}-based similarity measure, a single-source selection experiment
is conducted using the zero-shot transfer matrix in \cref{fig:eval_matrix}. For each target dataset, one source dataset
is selected according to each selection rule. The oracle source is defined as the candidate source achieving the lowest
minADE$_6$ on the target dataset.

The quality of a selected source is evaluated using four metrics. First, its rank is determined by ordering all
candidate sources according to their actual minADE$_6$ on the target dataset. Second, the top-3 rate measures the
fraction of targets for which the selected source is among the three best-performing candidate sources. Third, the
absolute performance gap is computed as the difference between the minADE$_6$ obtained using the selected source and
that obtained using the oracle source. Finally, the relative gap normalizes this difference by the oracle minADE$_6$.
The reported values are averaged across the target datasets.

As shown in \cref{tab:selection}, the \gls{KL}-based rule selects one of the three best-performing sources for $83.3\%$
of the target datasets and achieves a mean selected-source rank of $3.00$. It also produces the smallest mean absolute
and relative minADE$_6$ gaps, outperforming selection based on mean speed, source-dataset size, and random sampling.
These results indicate that the proposed similarity measure can provide a useful criterion for selecting a source
dataset when target-domain performance measurements are unavailable.

The experiment is restricted to single-source zero-shot transfer and does not consider multi-source training or
target-domain fine-tuning. The effectiveness of the proposed measure in these more general and complex transfer settings
therefore remains to be investigated. 

\input{tab/transfer_selection.tex}

\subsection{Covariance estimation ablation}

The covariance estimation ablation considers two design choices: the source of the covariance matrix and the estimator
applied to it. The covariance source is either the within-scene covariance, the between-scene covariance, or their sum,
with the latter defined in \eqref{eq:total_covariance}. The evaluated estimators range from the raw empirical covariance
to regularized and low-rank approximations.

\paragraph{Covariance estimators.}

Consider the decomposition of the empirical covariance matrix,
\[
    \boldsymbol{\Sigma}
    =
    \mathbf{U}\boldsymbol{\Lambda}\mathbf{U}^{\mathsf T},
\]
where $\mathbf{U}$ contains the eigenvectors and
$\boldsymbol{\Lambda}
=\operatorname{diag}(\lambda_1,\ldots,\lambda_d)$ contains the corresponding
eigenvalues. The four covariance estimators are defined as
\begin{alignat}{2}
&\widehat{\boldsymbol{\Sigma}}_{\mathrm{raw}}
    \quad &&= \boldsymbol{\Sigma},
\\
&\widehat{\boldsymbol{\Sigma}}_{\mathrm{jitter}}
    \quad &&= \boldsymbol{\Sigma} + \frac{\mathrm{tr}(\boldsymbol{\Sigma})}{L}\mathbf{I},
\\
&\widehat{\boldsymbol{\Sigma}}_{\mathrm{floor}}
    \quad &&= \mathbf{U}
    \operatorname{diag}\!\bigl(
        \max\{\lambda_1,\alpha\},
        \ldots,
        \max\{\lambda_d,\alpha\}
    \bigr)
    \mathbf{U}^{\mathsf T},
\\
&\widehat{\boldsymbol{\Sigma}}_{\mathrm{LR}}
    \quad &&= \mathbf{U}_r
    \operatorname{diag}(\lambda_1,\ldots,\lambda_r)
    \mathbf{U}_r^{\mathsf T}
    + \alpha \frac{\mathrm{tr}(\boldsymbol{\Sigma})}{L} \mathbf{I},
\end{alignat}
where $r$ denotes the retained rank and $\alpha$ controls the regularization
strength.

Performance was assessed primarily using Spearman's rank correlation coefficient, while the median KL divergence was
used as a secondary diagnostic of distributional discrepancy and sensitivity to covariance regularization. For each
latent dimension, the selected configuration was defined as the one with the lowest median KL among those
retaining at least 99\% of the maximum correlation.

\Cref{tab:covariance_ablation} compares the maximum-correlation and selected configurations. All optima use a low-rank
estimate of the summed covariance source. The selected configurations preserve near-maximal correlation while generally
exhibiting substantially lower median and upper-tail KL divergence. At 32 dimensions, the selected configuration
retains 99.3\% of the maximum correlation while reducing the median and 95th-percentile KL divergences by factors of
approximately 875 and 596, respectively.

\input{tab/covar_ablation.tex}

\begin{figure}[t]
    \centering
    \includegraphics[width=0.62\columnwidth]
        {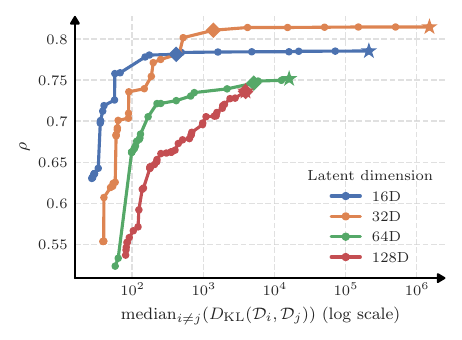}
    \vspace{-0.3cm}
    \caption{Relationship between Spearman's rank correlation coefficient $\rho$ and median KL divergence across the
evaluated covariance sources, estimators, ranks, and regularization strengths. The displayed frontiers are nondominated
under the reporting convention of maximizing $\rho$ and minimizing KL. Stars denote maximum-correlation configurations,
while diamonds denote the selected configurations.}
    \label{fig:kl_rho_pareto}
\end{figure}

\Cref{fig:kl_rho_pareto} shows the relationship between correlation and median KL under the selection convention of
maximizing correlation while preferring lower KL among near-optimal configurations. Each frontier contains
configurations that are nondominated within the same latent dimension; moving along a frontier changes the balance
between correlation and KL. The 32-dimensional frontier extends to the highest correlation and contains a strongly
regularized solution close to its maximum-correlation endpoint. Specifically, the selected configuration achieves
$\rho=0.811$, compared with the maximum value of 0.817, while reducing the median KL by approximately $875\times$.

Overall, the ablation reveals a clear interaction between covariance estimation and latent dimension. The best
correlation is obtained at 32 dimensions, while both lower and higher-dimensional representations perform worse. This
non-monotonic trend suggests that 32 dimensions provide a favorable balance between representational capacity and
reliable covariance estimation. Since the selected 32-dimensional configuration preserves 99.3\% of the maximum
correlation while avoiding the extreme KL values observed for the weakly regularized estimate, it was adopted as the
main covariance configuration.

%% file: tab/qcnet_params.tex
\begin{table}[b]
  \centering
  \caption{Modified hyperparameters of the QCNet trajectory prediction model.}
  \setlength{\tabcolsep}{12pt}
  \begin{tabular}{l c c}
    \toprule
    \textbf{Hyperparameter} & \textbf{Original}$^{1}$& \textbf{Modified} \\
    \midrule
    Hidden dimension               & $128$         & $48$           \\
    Number of heads                & $8$           & $6$            \\
    Head dimension                 & $16$          & $12$           \\
    Frequency bands                & $64$          & $48$           \\
    Number of parameters           & $7.66$M        & $1.35$M       \\
    \bottomrule
    \addlinespace[3pt]
    \multicolumn{3}{l}{$^{1}$\footnotesize{As reported in the original QCNet paper~\cite{zhou2023query} (Argoverse 2 version).}} \\
  \end{tabular}
  
    \label{tab:qcnet-hyperparams}
\end{table}

%% file: tab/qcnet_training.tex
\begin{table*}[t]
  \centering
  \caption{Dataset-specific training configurations for the QCNet trajectory
  prediction model. Batch sizes are determined by memory constraints, and
  learning rates are selected based on preliminary experiments to ensure
  stable convergence.}
  \label{tab:qcnet-training-config}

  \setlength{\tabcolsep}{3.2pt}
  \scriptsize
  \begin{tabular}{
    l
    c
    S[scientific-notation=true]
    c
    @{\hspace{6pt}}
    !{\color{lightgray}\vrule width 0.6pt}
    @{\hspace{6pt}}
    l
    c
    S[scientific-notation=true]
    c
  }
    \toprule
    \textbf{Dataset}
      & \textbf{Batch}
      & {\textbf{LR}}
      & \textbf{Epochs}
      &
    \textbf{Dataset}
      & \textbf{Batch}
      & {\textbf{LR}}
      & \textbf{Epochs} \\
    \midrule

    A43
      & 46
      & \num{7.5e-05}
      & 41
      &
    Lyft
      & 24
      & \num{5e-04}
      & 33 \\

    AD4CHE
      & 8
      & \num{5e-04}
      & 50
      &
    nuScenes
      & 60
      & \num{1e-04}
      & 16 \\

    ApolloScape
      & 34
      & \num{5e-04}
      & 12
      &
    openDD
      & 36
      & \num{7.5e-05}
      & 16 \\

    Argoverse
      & 16
      & \num{1e-04}
      & 26
      &
    rounD
      & 200
      & \num{5e-03}
      & 50 \\

    Argoverse 2
      & 8
      & \num{1e-04}
      & 20
      &
    SIND
      & 12
      & \num{5e-04}
      & 20 \\

    eth
      & 54
      & \num{1e-04}
      & 26
      &
    uniD
      & 55
      & \num{1e-04}
      & 24 \\

    exiD
      & 96
      & \num{1.25e-04}
      & 29
      &
    univ
      & 128
      & \num{1e-04}
      & 50 \\

    highD
      & 64
      & \num{1e-04}
      & 50
      &
    US-101
      & 8
      & \num{5e-04}
      & 20 \\

    hotel
      & 54
      & \num{1e-04}
      & 40
      &
    View-of-Delft
      & 64
      & \num{5e-04}
      & 50 \\

    I-80
      & 9
      & \num{5e-05}
      & 22
      &
    WOMD
      & 42
      & \num{1e-04}
      & 37 \\

    inD
      & 26
      & \num{7.5e-05}
      & 34
      &
    zara1
      & 54
      & \num{1e-04}
      & 50 \\

    INTERACTION
      & 64
      & \num{1e-04}
      & 50
      &
    zara2
      & 54
      & \num{1e-04}
      & 50 \\

    \bottomrule
  \end{tabular}
\end{table*}

%% file: tab/transfer_selection.tex
\begin{table}[t]
    \scriptsize
    \centering
    \renewcommand{\arraystretch}{1.15}
    \setlength{\tabcolsep}{5.3pt}
    \caption{
    Single-source selection performance averaged across target datasets.
    Rank and top-3 rate are based on the achieved minADE$_6$ of the selected
    source. The absolute and relative gaps are measured with respect to the
    oracle source.
}
    \label{tab:selection}
    \begin{tabular}{lcccc}
        \toprule
        Selection rule
        & \begin{tabular}[c]{@{}c@{}}
            Mean selected\\source rank $\downarrow$
          \end{tabular}
        & \begin{tabular}[c]{@{}c@{}}
            Selected source\\in top 3 $\uparrow$
          \end{tabular}
        & \begin{tabular}[c]{@{}c@{}}
            Mean absolute\\minADE$_6$ gap $\downarrow$
          \end{tabular}
        & \begin{tabular}[c]{@{}c@{}}
            Mean relative\\minADE$_6$ gap $\downarrow$
          \end{tabular} \\
        \midrule
        KL divergence
        & \textbf{3.00}
        & \textbf{83.3\%}
        & \textbf{0.100}
        & \textbf{21.0\%} \\

        Closest mean speed
        & 6.29
        & 37.5\%
        & 0.335
        & 69.5\% \\

        Largest source dataset
        & 9.08
        & 20.8\%
        & 0.984
        & 243.2\% \\

        Random selection
        & 11.9
        & 13.2\%
        & 2.59
        & 667.4\% \\
        \bottomrule
    \end{tabular}
\end{table}

%% file: tab/covar_ablation.tex
\begin{table}[t]
    \centering
    \caption{Maximum-correlation and selected covariance configurations.
    The selected configuration minimizes the median KL divergence among
    candidates retaining at least 99\% of the maximum $\rho$ for the
    corresponding latent dimension. LR denotes low-rank covariance, and Sum
    denotes the summed covariance source.}
    \label{tab:covariance_ablation}

    \scriptsize
    \setlength{\tabcolsep}{4pt}
    \renewcommand{\arraystretch}{1.1}
    \sisetup{detect-weight=true, detect-inline-weight=math}

    \begin{tabular}{
        c
        l
        l
        c
        c
        S[table-format=1.4]
        S[scientific-notation=true, table-format=1.2e2]
        S[scientific-notation=true, table-format=1.2e2]
    }
        \toprule
        \makecell{Latent\\dim.} &
        Candidate &
        \makecell{Est. /\\source} &
        Rank &
        $\alpha$ &
        {$\rho$} &
        {\makecell{KL\\median}} &
        {\makecell{KL\\p95}} \\
        \midrule

        \multirow{2}{*}{16}
        & Max.
        & LR / Sum
        & 8
        & $1{\times}10^{-3}$
        & 0.781
        & 4.15e2
        & 7.10e4 \\

        & Selected
        & LR / Sum
        & 8
        & $1{\times}10^{-3}$
        & 0.781
        & 4.15e2
        & 7.10e4 \\

        \addlinespace[2pt]

        \multirow{2}{*}{32}
        & Max.
        & LR / Sum
        & 16
        & $1{\times}10^{-7}$
        & 0.817
        & 1.21e6
        & 5.71e7 \\

        & \textbf{Selected}
        & \textbf{LR / Sum}
        & \textbf{16}
        & $\mathbf{1{\times}10^{-4}}$
        & \bfseries 0.811
        & \bfseries 1.38e3
        & \bfseries 9.58e4 \\

        \addlinespace[2pt]

        \multirow{2}{*}{64}
        & Max.
        & LR / Sum
        & 48
        & $1{\times}10^{-7}$
        & 0.758
        & 1.25e4
        & 7.02e5 \\

        & Selected
        & LR / Sum
        & 48
        & $3{\times}10^{-7}$
        & 0.753
        & 5.12e3
        & 3.16e5 \\

        \addlinespace[2pt]

        \multirow{2}{*}{128}
        & Max.
        & LR / Sum
        & 64
        & $1{\times}10^{-7}$
        & 0.736
        & 3.92e3
        & 5.42e5 \\

        & Selected
        & LR / Sum
        & 64
        & $1{\times}10^{-7}$
        & 0.736
        & 3.92e3
        & 5.42e5 \\

        \bottomrule
    \end{tabular}
\end{table}